# Scalable Bilinear $\pi$ Learning Using State and Action Features

Yichen Chen[*]   Lihong Li[†]   Mengdi Wang[‡]

**Abstract**

Approximate linear programming (ALP) represents one of the major algorithmic families to solve large-scale Markov decision processes (MDP). In this work, we study a primal-dual formulation of the ALP, and develop a scalable, model-free algorithm called *bilinear $\pi$ learning* for reinforcement learning when a sampling oracle is provided. This algorithm enjoys a number of advantages. First, it adopts (bi)linear models to represent the high-dimensional value function and state-action distributions, using given state and action features. Its run-time complexity depends on the number of features, not the size of the underlying MDPs. Second, it operates in a fully online fashion without having to store any sample, thus having minimal memory footprint. Third, we prove that it is sample-efficient, solving for the optimal policy to high precision with a sample complexity linear in the dimension of the parameter space.

## 1 Introduction

Reinforcement learning lies at the intersection between control, machine learning, and stochastic processes (Bertsekas and Tsitsiklis, 1996; Sutton and Barto, 1998). The objective is to learn an optimal policy of a controlled system from interaction data. The most studied model for a controlled stochastic system is the Markov decision process (MDP), i.e., a controlled random walk over a state space $\mathcal{S}$, where in each state $s \in \mathcal{S}$ one can choose an action $a$ from an action space $\mathcal{A}$ so that the random walk transitions to another state $s' \in \mathcal{S}$ with probability $P_a(s, s')$. In this paper, we do not assume the MDP model is explicitly known, but consider the setting where a *generative model* is given (see, e.g., Azar et al. (2013)). In other words, there is an oracle that takes $(s, a)$ as input and outputs a random $s'$ with probability $P_a(s, s')$. This is also known as the simulator-defined MDP in some literatures (Dietterich et al., 2013; Taleghan et al., 2015). Our goal is to find an optimal policy which, when runs on the MDP to generate an infinitely long trajectory, yields the highest average per-step reward in the limit.

Here, we focus on problems where the state and action spaces $\mathcal{S}$ and $\mathcal{A}$ are too large to be enumerated. In practice, it might be computationally challenging to even store a single state of the process (e.g., states could correspond to high-resolution images). Suppose that we are given a collection of state features $\phi : \mathcal{S} \mapsto \mathbb{R}^D$ and action features $\psi : \mathcal{A} \mapsto \mathbb{R}^U$. They map each state $s \in \mathcal{S}$ and action $a \in \mathcal{A}$ to column vectors $\phi(s) = (\phi_1(s), \ldots, \phi_D(s))^T$ and $\psi(a) = (\psi_1(a), \ldots, \psi_U(a))^T$, respectively, where $D \ll S := |\mathcal{S}|$ and $U \ll A := |\mathcal{A}|$.

Our primary interest is to develop a sample-efficient and computationally scalable algorithm, which takes advantage of the given features to solve an MDP with very large state and action spaces. Given the feature maps, $\phi$ and $\psi$, we adopt linear and bilinear models for approximating both the value function and the stationary state-action distribution of the MDP. By doing so, we can represent the value functions and state-action distributions, which are high-dimensional quantities, using a much smaller number of parameters.

---

[*]Department of Computer Science, Princeton University, Princeton, NJ, USA; e-mail: `yichenc@princeton.edu`

[†]Google Inc., Kirkland, WA, USA; e-mail: `lihongli.cs@gmail.com`

[‡]Department of Operations Research and Financial Engineering, Princeton University, Princeton, NJ, USA; e-mail: `mengdiw@princeton.edu`

**Contributions.** Our main contribution is a tractable, model-free primal-dual $\pi$-learning algorithm for such compact parametric representations. It incrementally updates parameters as new transitions are observed. With given state and action features, its per-iteration complexity is low, depending on $U$ and $D$, not $|\mathcal{S}|$ or $|\mathcal{A}|$:

- The new algorithm is inspired by a saddle-point formulation of policy optimization in MDPs, which we refer to as the *Bellman saddle-point problem*. We show a strong relation between the parametric saddle point problem and the original Bellman equation. In particular, the difference between solutions to these two problems can be quantified using the $\ell_\infty$- and $\ell_1$-errors of the parametric function classes that are used to approximate the optimal value function and state-action distribution respectively. In the special case where the approximation error is zero (also known as the "realizable" scenario), solving the Bellman saddle-point problem is equivalent to solving the original Bellman equation.
- Each iteration of the algorithm can be viewed as a stochastic primal-dual iteration for solving the Bellman saddle-point problem, where the value and policy updates are coupled together in light of strong duality. We study the sample complexity of the $\pi$ learning algorithm by analyzing the coupled primal-dual convergence process. We show that finding an $\epsilon$-optimal policy (comparing to the best approximate policy) requires a sample size that is linear with respect to $\frac{DU}{\epsilon^2}$. The sample complexity depends only on the numbers of state and action features. It is invariant with respect to the actual sizes of the state and action spaces.

**Notations.** The following notations are used throughout the paper. For any integer $n$, we use $[n]$ to denote the set of integers $\{1, 2, \ldots, n\}$. For a matrix $\Phi$ of size $m \times n$ and $1 \leq p \leq \infty$, the matrix $p$-norm is defined as $\|\Phi\|_p = \max\{\|\Phi v\|_p : v \in \mathbb{R}^n \text{ with } \|v\|_p = 1\}$. The $(a,b)$-norm of $\Phi$, denoted by $\|\Phi\|_{a,b}$, is defined as the $\ell_b$-norm of the column vector that consists of $\ell_a$-norm of the rows of $\Phi$. We use $\Phi_{i*}$ and $\Phi_{*j}$ to denote the $i$-th row and the $j$-th column of $\Phi$ respectively. For a vector $v$, we denote by $\text{diag}(v)$ the diagonal matrix with $v$ on its diagonal. We denote by $\mathbf{1}$ the all-one column vector. For two probability distributions $u, w$ over a finite set $X$, we denote by $D_{KL}(u\|w)$ their Kullback-Leibler divergence, i.e., $D_{KL}(u\|w) = \sum_{x \in X} u(x) \log \frac{u(x)}{w(x)}$. For two functions $f(x)$ and $g(x)$, we say that $f(x) = \mathcal{O}(g(x))$ if there exists a constant $C$ such that $|f(x)| \leq Cg(x)$ for all $x$. We use $\tilde{\mathcal{O}}(\cdot)$ as a variant of $\mathcal{O}(\cdot)$ that hides the logarithmic factors.

## 2 Preliminaries

We review the basics of infinite-horizon MDP with the average-reward criterion.

### 2.1 Infinite-Horizon Average-Reward MDP

Most of the paper focuses on the infinite-horizon average-reward Markov Decisions Problem (MDP), in which one aims to make an infinite sequence of decisions and optimize the average-per-time-step reward. An instance of the MDP can be described by a tuple $\mathcal{M} = (\mathcal{S}, \mathcal{A}, \mathcal{P}, \mathbf{r})$, where $\mathcal{S}$ is a state space, $\mathcal{A}$ is an action space, $\mathcal{P}$ is the collection of state-to-state transition probabilities $\mathcal{P} = \{P_a(s, s') \mid s, s' \in \mathcal{S}, a \in \mathcal{A}\}$, $\mathbf{r}$ is the collection of immediate reward functions $\mathbf{r} = \{r_a(s) \mid s \in \mathcal{S}, a \in \mathcal{A}\}$ with $r_a(s) \in [0, 1]$. In every step of the decision process, the system is in some state $s$, and an action $a$ is chosen by a control policy. The system then transitions to a next-state $s'$ with probability $P_a(s, s')$ with an immediate reward $r_a(s)$.

### 2.2 Policy Optimization

A randomized stationary policy can be represented by a matrix $\pi \in \Pi \subset \mathbb{R}^{S \times A}$, where $\Pi$ consists of non-negative matrices whose $s$-th row, denoted by $\pi_{s*}$, corresponds to a probability distribution over $\mathcal{A}$ for state $s$. The policy optimization problem is to maximize the infinite-horizon average reward over stationary policies:

$$\max_{\pi \in \Pi} \left\{ \bar{v}^\pi = \lim_{T \to \infty} \mathbf{E}^\pi \left[ \frac{1}{T} \sum_{t=1}^T r_{a_t}(s_t) \right] \right\}, \tag{1}$$

where $(s_1, a_1, s_2, a_2, \ldots, s_t, a_t, \ldots)$ are state-action transitions generated by the Markov decision process under $\pi$ from an arbitrary initial distribution, and the expectation $\mathbf{E}^\pi[\cdot]$ is taken over the entire trajectory.

We denote by $P^\pi$ the transition probability matrix of the MDP under a fixed policy $\pi$, i.e., $P^\pi(s, s') = \sum_{a \in \mathcal{A}} \pi_{s,a} P_a(s, s')$ for all $s, s' \in \mathcal{S}$. Note that the policy optimization problem (1) is equivalent to the following (Puterman, 2014)

$$\max_{\pi \in \Pi, \xi \in \mathbb{R}^\mathcal{S}} \sum_{s,a} \xi_s \pi_{s,a} r_a(s)$$
$$\text{subject to } \xi^\top P^\pi = \xi^\top, \xi \geq 0, \xi^\top \mathbf{1} = 1,$$

where the constraints require that $\xi$ is the stationary distribution of states under $P^\pi$, and the objective is an explicit expression of the average reward $\bar{v}^\pi$. By defining $\mu_{s,a} = \xi_s \pi_{s,a}$, the policy optimization becomes a linear program

$$\max_{\mu \in \mathbb{R}^{\mathcal{S} \times \mathcal{A}}} \sum_{s,a} \mu_{s,a} r_a(s)$$
$$\text{s.t. } \sum_a \mu_{*a}^\top P_a = \sum_a \mu_{*a}^\top, \mu \geq 0, \|\mu\|_{1,1} = 1, \tag{2}$$

where the constraint requires that $\mu$ is a *stationary state-action distribution* of the MDP under some policy. We denote by $\mu^*$ the optimal solution to (2). The optimal policy can be obtained by $\pi^*_{s,a} = \frac{\mu^*_{s,a}}{\|\mu^*_{s*}\|_1}$. As long as there exists an optimal solution, there exists an optimal basic solution (corresponding to a deterministic policy) (Puterman, 2014).

### 2.3 Bellman Equation

According to the theory of dynamic programming (Puterman, 2014; Bertsekas, 1995), the value $\bar{v}^*$ is the optimal average reward to the MDP $\mathcal{M}$ if and only if it satisfies the following *Bellman equation*:

$$\bar{v}^* + v^*_s = \max_{a \in \mathcal{A}} \left\{ \sum_{s' \in \mathcal{S}} P_a(s, s') v^*_{s'} + r_a(s) \right\}, \forall \, s \in \mathcal{S}, \tag{3}$$

for some vector $v^* \in \mathbb{R}^{|\mathcal{S}|}$. Here, $v^*$ is known as the difference-of-value vector, which we also refer to as the *value vector* or *value function* for short. Note that there exist multiple solutions of $v^*$ to (3) (by adding any constant shift), which do not affect our analysis. The results of the paper hold for any such $v^*$ unless stated otherwise. A stationary policy $\pi^*$ is an optimal policy of the MDP if it attains the elementwise maximization in the Bellman equation (Puterman, 2014, Theorem 8.4.5). The Bellman equation also admits an equivalent linear program:

$$\min_{\bar{v} \in \mathbb{R}, v \in \mathbb{R}^\mathcal{S}} \bar{v}$$
$$\text{subject to } \bar{v} \cdot \mathbf{1} + (I - P_a) v - r_a \geq 0, \quad \forall \, a \in \mathcal{A}. \tag{4}$$

It is well known that the linear program (4) is the dual of the policy optimization problem (2) (Puterman, 2014). Informally speaking, the Bellman equation and the policy optimization problem are dual to each other.

## 3 Bilinear Model Reduction of MDP using State and Action Features

In this section we describe a linear model for approximating the optimal value vector and a *bilinear* additive model for approximating the state-action distribution. Then we formulate a reduced-order primal-dual policy optimization problem, which we refer to as the Bellman saddle-point problem.

## 3.1 Using State and Action Features As Bilinear Bases

Suppose that we are given feature functions $\phi : \mathcal{S} \mapsto \mathbb{R}^D$ and $\psi : \mathcal{A} \mapsto \mathbb{R}^U$, which map the state space and action space into low-dimensional spaces, respectively. For simplicity of analysis, we assume the following:

**Assumption 1** (State and Action Features). *The state features $\phi_1(\cdot), \ldots, \phi_D(\cdot)$ and action features $\psi_1(\cdot), \ldots, \psi_U(\cdot)$ are probability density functions over the state space $\mathcal{S}$ and the action space $\mathcal{A}$, respectively. Besides, the state density functions and the action density functions are both linearly independent.*

Note that this assumption is not very strong: for finite state and action spaces, any nonnegative nonzero feature vector can be normalized to be a nonnegative vector whose entries sum to 1. Similarly, for general state and action spaces, any nonnegative and nonzero integrable feature functions can be normalized to become a probability density function. Such normalization is necessary for the convenience of analysis and sampling, which does not affect the scaling of our complexity bounds.

Let $(v^*, \pi^*)$ be a pair of optimal difference-of-value vector and optimal policy of the MDP. We adopt a (bi)linear model to approximate the value function and state-action distribution:

- The value function is approximated by a linear model, where we hope to find $\tilde{v} \in \mathbb{R}^D$ such that

$$v^*(\cdot) \approx \sum_{i=1}^{D} \tilde{v}_i \phi_i(\cdot).$$

- Representing the randomized policy $\pi$ is trickier, as it is a collection of conditional distributions. We will use an implicit representation. Specifically, we use a bilinear additive model to represent $\mu^*$, the stationary distribution of state-action pairs under the optimal policy:

$$\mu^*_{s,a} \approx \sum_{i=1}^{D} \sum_{u=1}^{U} \tilde{\mu}_{i,u} \phi_i(s) \psi_u(a),$$

where $\tilde{\mu} \in \mathbb{R}^{D \times U}$ is a matrix of parameters. The optimal policy is then approximated by

$$\pi^*_{s,a} \propto \sum_{i=1}^{D} \sum_{u=1}^{U} \tilde{\mu}_{i,u} \phi_i(s) \psi_u(a),$$

so that $\sum_a \pi^*_{s,a} = 1$ for all $s \in \mathcal{S}$.

**Remark.** Consider the case where $\mathcal{S}, \mathcal{A}$ are both large but finite sets. Let $\Phi \in \mathbb{R}^{|\mathcal{S}| \times D}, \Psi \in \mathbb{R}^{|\mathcal{A}| \times U}$ be the matrices consisting of column state features and column action features, respectively. We can write our models in matrix forms

$$v^* \approx \Phi \tilde{v}, \qquad \mu^* \approx \Phi \tilde{\mu} \Psi^\top.$$

In this case, Assumption 1 requires that the columns of $\Phi$ and $\Psi$ are nonnegative vectors whose entries sum to 1 and that $\Phi, \Psi$ are full rank.

## 3.2 Reduced-Order Bellman Saddle-Point Problem

The policy optimization problem (1) and the Bellman equation (3) admit linear programs (2) and (4), which are dual to each other. As suggested by Wang (2017), they can be equivalently formulated as a minimax problem:

$$\min_{v} \max_{\mu \geq 0, \|\mu\|_{1,1}=1} \sum_{a \in \mathcal{A}} \mu_{*a}^\top \big((P_a - I)v + r_a\big). \tag{5}$$

Its saddle point(s) coincide with the pair(s) of optimal value function and the corresponding state-action distribution. The optimal minimax value is equal to the optimal average reward $\bar{v}^* = \sum_{a \in \mathcal{A}} (\mu^*_{*a})^\top r_a$.

With the linear and bilinear models described in the previous subsection, we approximate the high-dimensional saddle point problem (5) with the following that involves lower-dimensional variables:

$$\min_{\tilde{v} \in \mathcal{V}} \max_{\tilde{\mu} \in \mathcal{U}} \sum_{a \in \mathcal{A}} \Psi_{a*} \tilde{\mu}^\top \Phi^\top ((P_a - I) \Phi \tilde{v} + r_a), \tag{6}$$

where $\mathcal{V}$ and $\mathcal{U}$ are certain primal and dual constraints to be specified later. Here, $\tilde{\mu}$ is of dimension $D \times U$. The primal and dual variables $(\tilde{v}, \tilde{\mu})$ are parameters of the bilinear models for the value function and state-action distribution. Equivalently, one can rewrite (6) into a sampling-friendly version:

$$\min_{\tilde{v} \in \mathcal{V}} \max_{\tilde{\mu} \in \mathcal{U}} \sum_{i=1}^{D} \sum_{u=1}^{U} \tilde{\mu}_{i,u} \mathbf{E}_{\phi_i, \psi_u} \left[ \phi_{s'}^T \tilde{v} - \phi_s^T \tilde{v} + r_a(s) \right], \tag{7}$$

where the expectation is taken over the random variables $(s, a, s')$ where $s \sim \phi_i$, $a \sim \psi_u$, $s' \sim P_a(s, \cdot)$.

### 3.3 Realizability of The Reduced Model

We introduce the following notion of realizability regarding the state and action features.

**Definition 1** (Realizability). *We say that the MDP $\mathcal{M}$ is realizable using state and action features $\phi, \psi$ if there exist $\tilde{v} \in \mathbb{R}^D$ and $\tilde{\mu} \in \mathbb{R}_+^{D \times U}$ so that $v^* = \Phi \tilde{v}$ and $\mu^* = \Phi \tilde{\mu} \Psi^\top$.*

We have the following result:

**Theorem 1.** *Let $\mathcal{M}$ be an MDP tuple that is realizable using $\phi, \psi$. Then there exists an optimal saddle point $(\tilde{v}^*, \tilde{\mu}^*)$ to problem (7) that satisfies*

$$v^* = \Phi \tilde{v}^*, \qquad \mu^* = \Phi \tilde{\mu}^* \Psi^\top.$$

The proof is straightforward by noting that (7) is a more restricted version of (5).

A natural question one may ask is what happens when the realizability condition does not hold. We wonder how well one can solve the high-dimensional MDP by solving the reduced-order saddle point problem (6) instead. We leave this question to Section 5.4, where we provide approximation guarantees for solving the misspecified saddle point problem, when realizability does not hold.

## 4 Bilinear $\pi$ Learning

This section develops the **bilinear $\pi$-learning** algorithm based on given state and action features. We discuss its implementation and run-time efficiency. Its sample efficiency is the subject of the next section.

### 4.1 The Algorithm

We propose to use a primal-dual algorithm to solve problem (6), which is given as in Algorithm 1. The algorithm makes updates to $\tilde{v}$ and $\tilde{\mu}$ while sampling state transitions. Let us denote by $(i_t, u_t, s_t, a_t, r_t, s'_t)$ the sample at iteration $t$. Let $\mathcal{U}, \mathcal{V}, M, \alpha, \beta$ be input parameters to be specified later. At the $t$-th iteration, the algorithm updates according to

$$\tilde{\mu}^{t+1} = \Pi_{\mathcal{U}, KL} \left( \frac{\tilde{\mu}^t \exp\left(\beta \Delta_\mu^{t+1}\right)}{\|\tilde{\mu}^t \exp\left(\beta \Delta_\mu^{t+1}\right)\|_{1,1}} \right), \qquad \tilde{v}^{t+1} = \Pi_\mathcal{V} \left( \tilde{v}^t - \alpha \Delta_v^{t+1} \right),$$

where $\Pi_{\mathcal{U},KL}(\mu) = \operatorname{argmin}_{\mu' \in \mathcal{U}} D_{KL}(\mu' \| \mu)$ denotes the information projection onto $\mathcal{U}$ with regard to the Kullback-Leibler divergence, $\Pi_{\mathcal{V}}(v) = \operatorname{argmin}_{v' \in \mathcal{V}} \|v - v'\|_2$ denotes the Euclidean projection onto $\mathcal{V}$, $\Delta_\mu^{t+1} \in \mathbb{R}^{D \times U}$ denotes the noisy gradient with respect to the dual variable, given by

$$\Delta_\mu^{t+1}(i, u) = \begin{cases} \frac{(\phi(s_t') - \phi(s_t))^\top \tilde{v}^t + r_t - M}{\tilde{\mu}_{i_t, u_t}^t}, & \text{if } (i, u) = (i_t, u_t) \\ 0, & \text{otherwise,} \end{cases}$$

and $\Delta_v^{t+1} \in \mathbb{R}^D$ denotes the noisy partial gradient with respect to the primal variable:

$$\Delta_v^{t+1} = \phi(s_t')^\top - \phi(s_t)^\top.$$

Essentially, each iteration is a stochastic primal-dual iteration (using $\|\cdot\|_2$ and KL divergence as the primal and dual Bregman divergence functions, respectively) for solving the saddle point problem. In particular, the dual update is related to the exponentiated gradient method. The algorithm has several interesting features that are worth noting:

- Upon obtaining a new sample, the $\pi$-learning algorithm makes multiplicative updates on dual variables, which resembles the exponentiated gradient methods used in bandit problems (Kivinen and Warmuth, 1997; Auer et al., 2002).
- The updates on dual variables involve projection with respect to the KL divergence. Similarly Bregmen divergence functions have been used for regularization for policy optimization (Schulman et al., 2015; Fox et al., 2016).
- The algorithm does not need to take the sample transitions as input, which can be high-dimensional quantities. Instead, it is sufficient to take as input the feature values $\phi(s), \psi(a)$ of the state and action, so that all the computation can be carried out in the low-dimensional parameter space.

---

**Algorithm 1** Bilinear $\pi$ Learning (Average Reward)

1: **Input:** The number of iterations $T > 0$, $\phi$, $\psi$.
2: **Input:** Stepsizes $\alpha$, $\beta$ and offset parameter $M$.
3: Set $\tilde{v}_i^1 = 0$, $\tilde{\mu}_{i,u}^1 = \frac{1}{DU}, i \in [D], u \in [U]$
4: Let $\tilde{\mu} = \tilde{\mu}^1$
5: **for** $t = 1, 2, 3, \ldots, T$ **do**
6:     Sample $(i, u)$ according to distribution $\tilde{\mu}$
7:     Sample $s$ according to distribution $\phi_i$
8:     Sample $a$ according to distribution $\psi_u$
9:     Sample $s'$ using $(s, a)$ and the generative model
10:     $\tilde{v}^{t+1} = \Pi_{\mathcal{V}} \left( \tilde{v}^t - \alpha \left( \phi(s') - \phi(s) \right)^\top \right)$
11:     $\tilde{\mu}_{i,u} = \tilde{\mu}_{i,u} \cdot \exp \left\{ \beta \cdot \frac{(\phi(s') - \phi(s))^\top \tilde{v}^t + r_a(s) - M}{\tilde{\mu}_{i,u}} \right\}$
12:     $\tilde{\mu} = \Pi_{\mathcal{U},KL} \left( \frac{\tilde{\mu}}{\|\tilde{\mu}\|_{1,1}} \right)$
13:     $\tilde{\mu}^{t+1} = \tilde{\mu}$
14: **end for**
15: $\hat{\mu} = \frac{1}{T} \sum_{t=1}^T \tilde{\mu}^t$
16: Let $\hat{\pi}_{s,a} = \frac{\Phi_{s*} \hat{\mu} \Psi_{a*}^\top}{\sum_{b \in \mathcal{A}} \Phi_{s*} \hat{\mu} \Psi_{b*}^\top}$
17: **Ouput:** $\hat{\pi}$

---

## 4.2 Computational Efficiency

Algorithm 1 is model-free in the sense that it never attempts to estimate the transition probability model of the MDP. Instead, it makes direct updates to the parameters of the value function and state-action distribution. It is worth emphasizing that it outputs a policy, not a value function.

Scalability is a significant advantage of Algorithm 1. The algorithm uses low-dimensional memory. It maintains two variables, the value parameters $\tilde{v}$ and the policy parameters $\tilde{\mu}$. The memory size is $\mathcal{O}(DU)$, which is the dimension of the policy parameters and does not depend on the actual dimension of the MDP. Furthermore, the algorithm operates in a purely streaming mode: for example, in the case where the decision process is a sequence of images, the proposed algorithm does not need to store any past image. Each iteration of the algorithm makes sparse updates. When the sets $\mathcal{V}, \mathcal{U}$ are of simple forms (see the next section for examples), each iteration can be implemented in runtime that is polynomial with respect to $D, U$.

## 5 Sample Complexity

We now turn to the sample-complexity analysis of Algorithm 1. We first analyze the convergence of of an inexact duality gap associated with the primal and dual iterates. Then we use the duality gap bound to derive the number of samples needed to find an $\epsilon$-optimal policy, provided that the realizability condition holds. Finally, we extend the analysis to the unrealizable case where the approximation error may be nonzero.

For simplicity of exposition, we focus on the case where the state and action spaces, $\mathcal{S}$ and $\mathcal{A}$, are finite. We will show that our sample complexity bounds do not scale with the sizes of the state and action spaces.

### 5.1 Assumptions

We make the following assumptions in the rest of the paper. They require that the state-action distributions of the MDP are within certain ranges and that the process is rapidly mixing. For convenience, we define $S = |\mathcal{S}|$ and $A = |\mathcal{A}|$.

**Assumption 2** (Uniformly Bounded Ergodicity). *The Markov decision process is ergodic under any stationary policy $\pi$, and there exists $\tau > 1$ such that*

$$\frac{1}{\sqrt{\tau}S} \cdot \mathbf{1} \leq \nu^\pi \leq \frac{\sqrt{\tau}}{S} \cdot \mathbf{1},$$

*where $\nu^\pi$ is the stationary distribution over the state space of the MDP under the policy $\pi$.*

This assumption is also used by Wang (2017). It implies that the MDP is unichain (Puterman, 2014).

**Assumption 3** (Fast Mixing Time). *There exists a finite time $t_{mix}$ such that for any stationary policy $\pi$*

$$t_{mix} \geq \min_t \left\{ t \ \Big| \ \|(P^\pi)^t(s, \cdot) - \nu^\pi\|_{\mathrm{TV}} \leq \frac{1}{4}, \forall s \in \mathcal{S} \right\},$$

*where $\|\cdot\|_{\mathrm{TV}}$ stands for the total variation.*

Under this assumption, there exists an optimal difference-of-value vector satisfying $\|v^*\|_\infty \leq 2t_{mix}$, according to Lemma 1 in (Wang, 2017). In the rest of the analysis, we focus on such $v^*$.

In what follows, we choose the primal constraint $\mathcal{V}$ as

$$\mathcal{V} = \{\tilde{v} \in \mathbb{R}^D \big| \|\Phi\tilde{v}\|_\infty \leq 2t_{mix}\},$$

and the dual constraint $\mathcal{U}$ as

$$\mathcal{U} = \left\{ \tilde{\mu} \ \Big| \ \tilde{\mu} \geq 0, \|\tilde{\mu}\|_{1,1} = 1, \sum_{a \in \mathcal{A}} \Phi\tilde{\mu}\Psi^\top \geq \frac{1}{\sqrt{\tau}S} \right\}.$$

Note that $\mathcal{U}$ is guaranteed to be nonempty as long as a constant-valued state feature function is included.

In the algorithm, we choose the parameters as

$$\beta = \frac{1}{5t_{mix}}\sqrt{\frac{\log(DU)}{TDU}}, \qquad \alpha = \frac{t_{mix}}{\lambda_{\min}(\Phi^\top\Phi)\|\Phi\|_{2,\infty}}\sqrt{\frac{D}{T}}, \qquad M = 4t_{mix} + 1. \tag{8}$$

## 5.2 Primal-Dual Convergence

In this section, we analyze the convergence of Algorithm 1. It can be viewed as a stochastic approximation algorithm, in which we use only a single sample per iteration.

Let $(\check{v}, \check{\mu})$ be the best approximation to $(v^*, \mu^*)$ using the linear models and given features $\phi, \psi$:

$$\check{\mu} = \mathrm{argmin}_{\tilde{\mu} \in \mathcal{U}} \|\Phi \tilde{\mu} \Psi^\top - \mu^*\|_{1,1}, \qquad \check{v} = \mathrm{argmin}_{\tilde{v} \in \mathcal{V}} \|\Phi \tilde{v} - v^*\|_\infty.$$

Our first result concerns the convergence of an inexact duality gap that is associated with the primal-dual iterates.

**Theorem 2 (Inexact Duality Gap).** *Let $\mathcal{M} = (\mathcal{S}, \mathcal{A}, \mathcal{P}, \mathbf{r})$ be an arbitrary MDP satisfying Assumptions 2 and 3. Let $\Phi \in \mathbb{R}^{S \times D}$ and $\Psi \in \mathbb{R}^{A \times U}$ satisfy Assumption 1. Then the sequence of iterates $\{\tilde{\mu}^t, \tilde{v}^t\}_{t=1}^T$ generated by Algorithm 1 satisfies the following:*

$$\sum_{a \in \mathcal{A}} r_a^\top \Phi \check{\mu} \Psi_{a*}^\top + \frac{1}{T} \sum_{t=1}^T \mathbf{E}\Big[\sum_{a \in \mathcal{A}}((I - P_a)\Phi\check{v} - r_a)^\top \Phi\tilde{\mu}^t \Psi_{a*}^\top - \sum_{a \in \mathcal{A}} (\Phi\check{\mu}\Psi_{a*}^\top)^\top (I - P_a) \Phi \tilde{v}^t\Big] \quad (9)$$
$$= \tilde{\mathcal{O}}\left(t_{mix}\left(c_\Phi + \sqrt{U \log(DU)}\right) \sqrt{\frac{D}{T}}\right),$$

*where $c_\Phi = \frac{\|\Phi\|_{2,\infty}}{\lambda_{\min}(\Phi^\top \Phi)}$ is a constant, and $\lambda_{\min}(\Phi^\top \Phi)$ is the smallest nonnegative eigenvalue of $\Phi^\top \Phi$.*

The constant $c_\Phi = \frac{\|\Phi\|_{2,\infty}}{\lambda_{\min}(\Phi^\top \Phi)}$ is feature-dependent, but it does *not* scale with $S$. Recall that each column of $\Phi$ is a nonnegative vector that sums to 1, therefore $\|\Phi\|_{2,\infty}$ and $\lambda_{\min}(\Phi^\top \Phi)$ do not scale with $S$. The constant $c_\Phi$ would be finite and dimension-free as long as $\lambda_{\min}(\Phi^\top \Phi)$ is bounded from below - a form of "restricted eigenvalue condition" that is commonly used for analysis of linear models.

To prove Theorem 2, we first establish recursions on the KL divergence $D_{KL}(\check{\mu}\|\tilde{\mu}^t) = \sum_{u=1}^U \sum_{i=1}^D \check{\mu}_{i,u} \log \frac{\check{\mu}_{i,u}}{\tilde{\mu}_{i,u}^t}$ and on the squared distance $\|\tilde{v}^t - \check{v}\|_2^2$ via the following two lemmas. We denote by $\mathcal{F}_t$ the collection of all the random variables revealed up to the $t$-th iteration of Algorithm 1.

**Lemma 1.** *Given $\mathcal{F}_t$, the KL divergence $D_{KL}(\check{\mu}\|\tilde{\mu}^{t+1})$ satisfies for $t \geq 1$ that*

$$\mathbf{E}[D_{KL}(\check{\mu}\|\tilde{\mu}^{t+1}) \mid \mathcal{F}_t] - D_{KL}(\check{\mu}\|\tilde{\mu}^t) \leq 50\beta^2 DU t_{mix}^2 + \beta \sum_{a \in \mathcal{A}} \Psi_{a*}(\tilde{\mu}^t - \check{\mu})^\top \Phi^\top ((P_a - I)\tilde{v}^t + r_a). \quad (10)$$

*Moreover, since $\tilde{\mu}_{i,u}^1 = \frac{1}{DU}$ for all $u \in [U], i \in [D]$ as in Algorithm 1, we have $D_{KL}(\check{\mu}\|\tilde{\mu}^1) \leq \log(DU)$.*

**Lemma 2.** *Given $\mathcal{F}_t$, the squared error $\|\tilde{v}^t - \check{v}\|_2^2$ satisfies*

$$\mathbf{E}[\|\tilde{v}^{t+1} - \check{v}\|_2^2 \mid \mathcal{F}_t] - \|\tilde{v}^t - \check{v}\|_2^2 \leq 2\alpha \sum_{a \in \mathcal{A}} \Psi_{a*}(\tilde{\mu}^t)^\top \Phi^\top (I - P_a) \Phi(\tilde{v}^t - \check{v}) + 4\alpha^2 \|\Phi\|_{2,\infty}^2, \quad (11)$$

*and $\|\tilde{v}^1 - \check{v}\|_2^2 \leq \frac{4D t_{mix}^2}{\lambda_{\min}^2(\Phi^\top \Phi)}$.*

The proofs of Lemma 1 and 2 are deferred to the appendix. Using these two lemmas, we prove Theorem 2 as follows.

*Proof of Theorem 2.* We first define $\mathcal{G}_t = \sum_{a \in \mathcal{A}} (\Phi \tilde{\mu}^t \Psi_{a*}^\top)^\top((I-P_a)\Phi\check{v} - r_a) + \sum_{a \in \mathcal{A}} r_a^\top \Phi \check{\mu} \Psi_{a*}^\top - \sum_{a \in \mathcal{A}} (\Phi \check{\mu} \Psi_{a*}^\top)^\top (I - P_a) \Phi \tilde{v}^t$. If we take the sum of equation (10) and (15), we obtain that

$$\mathcal{G}_t \leq \frac{D_{KL}(\check{\mu}\|\tilde{\mu}^t) - \mathbf{E}[D_{KL}(\check{\mu}\|\tilde{\mu}^{t+1}) \mid \mathcal{F}_t]}{\beta} + 50\beta DU t_{mix}^2 + \frac{\|\tilde{v}^t - \check{v}\|_2^2 - \mathbf{E}[\|\tilde{v}^{t+1} - \check{v}\|_2^2 \mid \mathcal{F}_t]}{2\alpha} + 2\alpha \|\Phi\|_{2,\infty}^2.$$

Applying the stepsizes in (8) and (8), taking the average over $t$, and using iterated expectations, we have

$$\sum_{t=1}^{T} \frac{\mathbf{E}[\mathcal{G}_t]}{T} \leq \frac{D_{KL}(\check{\mu}\|\tilde{\mu}^1)}{T\beta} + 50\beta DU t_{mix}^2 + \frac{\|\tilde{v}^1 - \check{v}\|_2^2}{2T\alpha} + 2\alpha \|\Phi\|_{2,\infty}^2$$

$$\leq 15 t_{mix} \sqrt{\frac{DU \log(DU)}{T}} + \frac{4 t_{mix} \|\Phi\|_{2,\infty}}{\lambda_{\min}(\Phi^\top \Phi)} \sqrt{\frac{D}{T}},$$

where the second inequality is due to Lemmas 1 and 2. Recall that $\frac{1}{T}\sum_{t=1}^{T} \mathbf{E}[\mathcal{G}_t]$ is exactly the LHS of equation (9). The proof is thus completed. □

### 5.3 Convergence To Optimal Policy: The Realizable Case

We now study the sample complexity of Algorithm 1 for obtaining a near-optimal policy in the realizable case. Recall that we are interested in finding an approximately-optimal policy which performs comparably to the optimal policy in the Markov decision process. Therefore, we wish to quantify the average reward for the policy learned by the bilinear $\pi$ learning algorithm:

**Theorem 3 (Convergence with Realizability).** *Let $\mathcal{M} = (\mathcal{S}, \mathcal{A}, \mathcal{P}, \mathbf{r})$ be an arbitrary MDP satisfying Assumptions 2 and 3. Let $\Phi \in \mathbb{R}^{S \times D}$ and $\Psi \in \mathbb{R}^{A \times U}$ satisfy Assumption 1. Suppose that the MDP $\mathcal{M}$ is realizable using state and action features $\Phi, \Psi$. Then, the policy $\hat{\pi}$ generated by Algorithm 1 after running $T$ time steps satisfies*

$$\bar{v}^* - \mathbf{E}[\bar{v}^{\hat{\pi}}] \leq \tilde{\mathcal{O}}\left(\tau t_{mix}\left(c_\Phi + \sqrt{U \log(DU)}\right)\sqrt{\frac{D}{T}}\right),$$

*where $\bar{v}^{\hat{\pi}}$ is the average reward of the policy $\hat{\pi}$.*

*Proof.* In the realizable case, we can write $\mu^* = \Phi \check{\mu} \Psi^\top$ and $v^* = \Phi \check{v}$. Note that $\mu^*$ is the stationary distribution under the optimal policy, which gives us that $\sum_{a \in \mathcal{A}} (\Phi \check{\mu} \Psi_{a*}^\top)^\top (I - P_a) = \sum_{a \in \mathcal{A}} (\mu_{*a}^*)^\top (I - P_a) = 0$, where $\mu_{*a}^*$ is the $a$-th column of $\mu^*$. It is also known that $\sum_{a \in \mathcal{A}} (\mu_{*a}^*)^\top r_a = \bar{v}^*$. We can simplify the result of Theorem 2 to

$$\bar{v}^* + \frac{1}{T}\sum_{t=1}^{T} \mathbf{E}\Big[\sum_{a \in \mathcal{A}} ((I - P_a)v^* - r_a)^\top \Phi \tilde{\mu}^t \Psi_{a*}^\top\Big] = \tilde{\mathcal{O}}\left(t_{mix}\left(c_\Phi + \sqrt{U \log(DU)}\right)\sqrt{\frac{D}{T}}\right), \quad (12)$$

where $\tilde{\mu}^t$ is the sequence of dual variables computed in Algorithm 1.

Recall that the probability of choosing action $a$ at state $s$, for the policy output by Algorithm 1, is

$$\hat{\pi}_{s,a} = \frac{\Phi_{s*}\hat{\mu}\Psi_{a*}^\top}{\sum_{a' \in \mathcal{A}} \Phi_{s*}\hat{\mu}\Psi_{a'*}^\top},$$

where $\hat{\mu} = \frac{1}{T}\sum_{t=1}^{T} \tilde{\mu}^t$. We let $\xi = (\xi_1, \ldots, \xi_S)$ be such that $\xi_s = \sum_{a' \in \mathcal{A}} \Phi_{s*}\hat{\mu}\Psi_{a'*}^\top$ for $s \in \mathcal{S}$. Denote by $\nu^{\hat{\pi}}$ the stationary distribution of the MDP under the policy $\hat{\pi}$. Using the definition of the average reward, we have

$$\bar{v}^* - \bar{v}^{\hat{\pi}} = \sum_{s \in \mathcal{S}} \sum_{a \in \mathcal{A}} \nu_s^{\hat{\pi}} \hat{\pi}_{s,a}(\bar{v}^* - r_a(s)) = \sum_{a \in \mathcal{A}} (\nu^{\hat{\pi}})^\top \mathrm{diag}(\hat{\pi}_{*a})((I - P_a)v^* + \bar{v}^* \cdot \mathbf{1} - r_a).,$$

where the last equality is because $\nu^{\hat{\pi}}$ is the stationary distribution and hence $\sum_{a \in \mathcal{A}} (\nu^{\hat{\pi}})^\top \mathrm{diag}(\hat{\pi}_{*a})(I - P_a) = \mathbf{0}$. From Assumption 2 and the definition of the dual constraint $\mathcal{U}$, we have $\nu_s^{\hat{\pi}} \leq \frac{\sqrt{\tau}}{S} = \tau \cdot \frac{1}{\sqrt{\tau}S} \leq \tau \xi_s$ for any state $s$. Also, we have $(I - P_a)v^* + \bar{v}^* \cdot \mathbf{1} - r_a \geq 0$ by the Bellman equation. This gives us

$$\bar{v}^* - \bar{v}^{\hat{\pi}} \leq \tau \sum_{a \in \mathcal{A}} \xi^\top \mathrm{diag}(\hat{\pi}_{*a})((I - P_a)v^* + \bar{v}^* \cdot \mathbf{1} - r_a).$$

By the definition of $\xi$, we have $\frac{1}{T}\sum_{t=1}^{T} \Phi_{s*} \tilde{\mu}^t \Psi_{a*}^\top = \xi_s \hat{\pi}_{s,a}$. Also note that $\sum_{a\in\mathcal{A}} \xi^\top \text{diag}(\hat{\pi}_{*a}) \cdot \mathbf{1} = 1$. It then follows from the last expression and equation (12) that

$$\bar{v}^* - \mathbf{E}[\bar{v}^{\hat{\pi}}] \leq \tau \left( \bar{v}^* + \frac{1}{T} \sum_{t=1}^{T} \mathbf{E}\left[ \sum_{a\in\mathcal{A}} ((I - P_a)v^* - r_a)^\top \Phi \tilde{\mu}^t \Psi_{a*}^\top \right] \right) \quad (13)$$
$$\leq \tilde{\mathcal{O}}\left( \tau t_{mix} \left( c_\Phi + \sqrt{U \log(DU)} \right) \sqrt{\frac{D}{T}} \right),$$

which completes the proof. $\square$

Theorem 3 suggests that, the bilinear $\pi$ learning algorithm achieves an $\epsilon$-optimal policy by using the sample size

$$\tilde{\mathcal{O}}\left( \frac{DU \log(DU)}{\epsilon^2} \right),$$

with other parameters fixed. This sample complexity bound depends linearly on the dimension of policy parameters. It is scale-free with respect to the sizes of the underlying MDP.

### 5.4 Convergence To Approximately Optimal Policy: The Unrealizable Case

We finally turn to the case where the realizability may not hold. The following theorem generalizes the results of Theorem 3 to the unrealizable case.

**Theorem 4 (Convergence without Realizability).** *Let $\mathcal{M} = (\mathcal{S}, \mathcal{A}, \mathcal{P}, \mathbf{r})$ be an arbitrary MDP satisfying Assumptions 2 and 3. Let $\Phi \in \mathbb{R}^{S \times D}$ and $\Psi \in \mathbb{R}^{A \times U}$ satisfy Assumption 1. Then the policy computed by Algorithm 1 after running $T$ time steps satisfies*

$$\bar{v}^* - \mathbf{E}[\bar{v}^{\hat{\pi}}] \leq \tilde{\mathcal{O}}\left( \tau t_{mix} \left( c_\Phi + \sqrt{U \log(DU)} \right) \sqrt{\frac{D}{T}} + \tau \cdot \min_{\tilde{v} \in \mathcal{V}} \|\Phi \tilde{v} - v^*\|_\infty + \tau t_{mix} \cdot \min_{\tilde{\mu} \in \mathcal{U}} \|\Phi \tilde{\mu} \Psi^\top - \mu^*\|_{1,1} \right),$$

*where $\bar{v}^{\hat{\pi}}$ is the average reward of the policy $\hat{\pi}$.*

The proof requires a careful characterization of the approximation error, and is deferred to the appendix. The approximation error has two parts. The first part $\min_{\tilde{v} \in \mathcal{V}} \|\Phi \tilde{v} - v^*\|_\infty$ is the $\ell_\infty$ distance from the optimal value function $v^*$ to the span of state featuers. The second part $\min_{\tilde{\mu} \in \mathcal{U}} \|\Phi \tilde{\mu} \Psi^\top - \mu^*\|_{1,1}$ is the total variation distance between the optimal state-action distribution and the best approximate distribution in the product space generated by state and action features.

## 6 Related Literatures

The primal-dual approach we developed in this paper is based on the well-known linear program (LP) formulation of the Bellman equation (d'Epenoux, 1963; Puterman, 2014). The exact LP has been extended to approximate linear programs (ALP) to tackle large-scale problems, by using compact representations of the value function. For example, Schweitzer and Seidmann (1985) and de Farias and Van Roy (2003) propose to represent the value function as a linear combination of basis functions, but the number of constraints depends on the number of state-action pairs, which motivated a lot of research on how to select a small subset of constraints without introducing much suboptimality in the reduced ALP (e.g., de Farias and Benjamin Van Roy (2004)). Others have studied various approaches to *compress* the large constraint set into a smaller one (Taylor and Parr, 2012; Lakshminarayanan et al., 2017). These prior works focus on the *quality* of the ALP solution—the distance it is from the optimal value function, but not so much on developing efficient algorithms to solve it. An exception is Abbasi-Yadkori et al. (2014), who reduces the ALP to stochastic convex optimization. However, their algorithm requires certain knowledge of the transition probabilities that is not easily available for all problems. In contrast, our work provides a concrete algorithm to solve the ALP when a bilinear representation is used, with strong guarantees on its computation and sample complexity; the algorithm only requires a sampling oracle.

A similar saddle-point approach has been studied by other authors. Chen and Wang (2016) and Wang (2017) propose provably fast solvers for finite MDPs, but those algorithms are not expected to scale well when the number of states/actions becomes large. Our algorithm's complexity depends on the number of basis functions, which has a much lower dimension than the number of states and actions. Dai et al. (2018) considers a related yet different primal-dual formulation that also allows compact representations, but no rate of convergence is provided. It should be noted that there is another line of research that studies saddle-point formulations that result from the fixed-point (as opposed to LP) view of the Bellman equation for fixed policy *evaluation* (Mahadevan et al., 2014; Macua et al., 2015; Du et al., 2017; Dai et al., 2017). In contrast, we consider policy optimization, which is substantially more challenging. It is an interesting future direction to study the connections between these two saddle-point formulations.

While our work is mostly related to the ALP family, another important class of methods to solve large-scale MDPs is approximate dynamic programming, or ADP (Bertsekas and Tsitsiklis, 1996). Linear basis functions have been used for both policy evaluation and optimization (Tsitsiklis and Van Roy, 1997; Nedić and Bertsekas, 2003; Lagoudakis and Parr, 2003; Melo et al., 2008; Sutton et al., 2009). However, except in certain special cases, ADP with linear approximation can be inherently unstable, leading to undesired situations including oscillation and even divergence. The Greedy-GQ is an interesting control algorithm with linear approximation that has convergence guarantees under a relatively mild assumption (Maei et al., 2010). But no finite-sample convergence is provided, and the algorithm requires two time-scale updates that can cause difficulties in practice. Our primal-dual algorithm not only is provably stable, but also enjoys strong finite-sample convergence rate.

Finally, our use of bilinear representations to compactly represent distributions over state-action pairs also appears new to the best of our knowledge. The most relevant model that has been studied seems to be (Elkan, 2011) which used a bilinear state-action representation for $Q$-functions. Previous work also uses linear approximation of value functions (Bertsekas and Tsitsiklis, 1996; Elkan, 2011), or state transition probabilities (Wang et al., 2007; Yang et al., 2009). In contrast, our representation allows one to work with large state and/or action spaces: the state and action are first mapped to their respective low-dimensional spaces, and a weight matrix is used to compute the distribution of that state-action pair. Such a technique may find use in other context when a distribution over state-action needs to be represented explicitly.

# 7 Summary

We provide a scalable model-free method, bilinear $\pi$ learning, for reinforcement learning, when a sampling oracle is provided. The method has its origin in a primal-dual formulation of the policy optimization problem. It adopts (bi)linear models to approximate the high-dimensional value functions and state-action distributions, using given state and action features. The approach enjoys a number of advantages. First, it is a compact method that has very small memory footprint and makes updates to low-dimensional variables. Its run-time and space complexities do not depend on the actual dimension of the MDP, so is scalable to large-scale reinforcement learning problems. Furthermore, the $\pi$-learning method is sample-efficient, which solves the approximate Bellman saddle-point problem to high precision using a small number of observations. The sample complexity of the compact $\pi$-learning method is linear in the dimension of the reduced parameter space, and again does not depend on the size of the underlying MDP.

We mention a few exciting directions for future research. First, how can we generalize our approach to nonlinear approximations such as neural networks? Second, how can we generalize the algorithm to online reinforcement learning, with the learner can only follow a single trajectory, with PAC (Strehl et al., 2009) or regret (Jaksch et al., 2010) guarantees? This will require explicitly addressing the exploration problem. Last but not least, how can we inject prior knowledge, such as sparsity (Kolter and Ng, 2009), in the primal-dual formulation and develop corresponding algorithms to take advantage of such information?

# References

ABBASI-YADKORI, Y., BARTLETT, P. L. and MALEK, A. (2014). Linear programming for large-scale Markov decision problems. In *Proceedings of the 31st International Conference on Machine Learning*.

AUER, P., CESA-BIANCHI, N., FREUND, Y. and SCHAPIRE, R. E. (2002). The nonstochastic multiarmed bandit problem. *SIAM Journal on Computing* **32** 48–77.

AZAR, M. G., MUNOS, R. and KAPPEN, H. J. (2013). Minimax PAC bounds on the sample complexity of reinforcement learning with a generative model. *Machine Learning* **91** 325–349.

BERTSEKAS, D. P. (1995). *Dynamic programming and optimal control*, vol. 1. Athena Scientific, Belmont, MA.

BERTSEKAS, D. P. and TSITSIKLIS, J. N. (1996). *Neuro-Dynamic Programming*. Athena Scientific.

CHEN, Y. and WANG, M. (2016). Stochastic primal-dual methods and sample complexity of reinforcement learning. *arXiv preprint arXiv:1612.02516* .

COVER, T. M. and THOMAS, J. A. (2012). *Elements of information theory*. John Wiley & Sons.

DAI, B., HE, N., PAN, Y., BOOTS, B. and SONG, L. (2017). Learning from conditional distributions via dual embeddings. In *Proceedings of the 20th International Conference on Artificial Intelligence and Statistics (AISTATS)*.

DAI, B., SHAW, A., HE, N., LI, L. and SONG, L. (2018). Boosting the actor with dual critic. In *Proceedings of the 6th International Conference on Learning Representations (ICLR)*. ArXiv:1712.10282.

DE FARIAS, D. P. and VAN ROY, B. (2003). The linear programming approach to approximate dynamic programming. *Operations Research* **51** 850–865.

DE FARIAS AND BENJAMIN VAN ROY, D. P. (2004). On constraint sampling in the linear programming approach to approximate dynamic programming. *Mathematics of Operations Research* **29** 462–478.

D'EPENOUX, F. (1963). A probabilistic production and inventory problem. *Management Science* **10** 98–108.

DIETTERICH, T. G., TALEGHAN, M. A. and CROWLEY, M. (2013). PAC optimal planning for invasive species management: Improved exploration for reinforcement learning from simulator-defined MDPs. In *AAAI*.

DU, S. S., CHEN, J., LI, L., XIAO, L. and ZHOU, D. (2017). Stochastic variance reduction methods for policy evaluation. In *Proceedings of the 34th International Conference on Machine Learning (ICML)*.

ELKAN, C. (2011). Reinforcement learning with a bilinear Q function. In *Recent Advances in Reinforcement Learning - 9th European Workshop (EWRL)*. No. 7188 in Lecture Notes in Computer Science.

FOX, R., PAKMAN, A. and TISHBY, N. (2016). Taming the noise in reinforcement learning via soft updates. In *Proceedings of the Thirty-Second Conference on Uncertainty in Artificial Intelligence (UAI)*.

JAKSCH, T., ORTNER, R. and AUER, P. (2010). Near-optimal regret bounds for reinforcement learning. *Journal of Machine Learning Research* **11** 1563–1600.

KIVINEN, J. and WARMUTH, M. K. (1997). Exponentiated gradient versus gradient descent for linear predictors. *Information and Computation* **132** 1–63.

KOLTER, J. Z. and NG, A. Y. (2009). Regularization and feature selection in least-squares temporal difference learning. In *Proceedings of the 26th International Conference on Machine Learning (ICML)*.

Lagoudakis, M. G. and Parr, R. (2003). Least-squares policy iteration. *Journal of Machine Learning Research* **4** 1107–1149.

Lakshminarayanan, C., Bhatnagar, S. and Szepesvári, C. (2017). A linearly relaxed approximate linear program for Markov decision processes. *IEEE Transactions on Automatic Control* **63** 1185–1191.

Macua, S. V., Chen, J., Zazo, S. and Sayed, A. H. (2015). Distributed policy evaluation under multiple behavior strategies. *IEEE Transactions on Automatic Control* **60** 1260–1274.

Maei, H. R., Szepesvri, C., Bhatnagar, S. and Sutton, R. S. (2010). Toward off-policy learning control with function approximation. In *Proceedings of the 27th International Conference on Machine Learning (ICML)*.

Mahadevan, S., Liu, B., Thomas, P. S., Dabney, W., Giguere, S., Jacek, N., Gemp, I. and Liu, J. (2014). Proximal reinforcement learning: A new theory of sequential decision making in primal-dual spaces. CoRR abs/1405.6757.

Melo, F. S., Meyn, S. P. and Ribeiro, M. I. (2008). An analysis of reinforcement learning with function approximation. In *Proceedings of the 25th International Conference on Machine Learning (ICML)*.

Nedić, A. and Bertsekas, D. (2003). Least squares policy evaluation algorithms with linear function approximation. *Discrete Event Dynamic Systems: Theory and Applications* **13** 79–110.

Puterman, M. L. (2014). *Markov decision processes: discrete stochastic dynamic programming.* John Wiley & Sons.

Schulman, J., Levine, S., Abbeel, P., Jordan, M. I. and Moritz, P. (2015). Trust region policy optimization. In *Proceedings of the Thirty-Second International Conference on Machine Learning (ICML)*.

Schweitzer, P. J. and Seidmann, A. (1985). Generalized polynomial approximations in Markovian decision processes. *Journal of Mathematical Analysis and Applications* **110** 568–582.

Strehl, A. L., Li, L. and Littman, M. L. (2009). Reinforcement learning in finite MDPs: PAC analysis. *Journal of Machine Learning Research* **10** 2413–2444.

Sutton, R. S. and Barto, A. G. (1998). *Reinforcement learning: An introduction.* MIT press.

Sutton, R. S., Szepesvári, C. and Maei, H. R. (2009). A convergent $o(n)$ algorithm for off-policy temporal-difference learning with linear function approximation. In *Advances in Neural Information Processing Systems 21 (NIPS)*.

Taleghan, M. A., Dietterich, T. G., Crowley, M., Hall, K. and Albers, H. J. (2015). PAC optimal mdp planning with application to invasive species management. *Journal of Machine Learning Research* **16** 3877–3903.

Taylor, G. and Parr, R. (2012). Value function approximation in noisy environments using locally smoothed regularized approximate linear programs. In *Proceedings of the 28th Conference on Uncertainty in Artificial Intelligence (UAI)*.

Tsitsiklis, J. N. and Van Roy, B. (1997). An analysis of temporal-difference learning with function approximation. *IEEE Transactions on Automatic Control* **42** 674–690.

Wang, M. (2017). Primal-dual $\pi$ learning: Sample complexity and sublinear run time for ergodic markov decision problems. *arXiv preprint arXiv:1710.06100* .

Wang, T., Lizotte, D. J., Bowling, M. H. and Schuurmans, D. (2007). Stable dual dynamic programming. In *Advances in Neural Information Processing Systems 20 (NIPS)*.

Yang, M., Li, Y. and Schuurmans, D. (2009). Dual temporal difference learning. In *Proceedings of the 12th International Conference on Artificial Intelligence and Statistics (AISTATS)*.

# A Proof of Lemma 1

*Proof.* Let us write $\tilde{\mu}^{t+1} = \Pi_{\mathcal{U},KL}(\tilde{\mu}^{t+1/2})$ where $\tilde{\mu}^{t+1/2}$ is the update vector prior to the projection step. Denote by $(i_t, u_t, s_t, a_t, s'_t, r_t)$ the sample at iteration $t$. Define the vector $\Delta^{t+1} \in \mathbb{R}^{D \times U}$ to be $\Delta_{i_t,u_t}^{t+1} = \frac{\Phi_{s'_t *}\tilde{v}^t - \Phi_{s_t *}\tilde{v}^t + r_t - M}{\tilde{\mu}_{i_t,u_t}^t}$ and $\Delta_{i,u}^{t+1} = 0$ for all $(i,u) \neq (i_t, u_t)$. Then the vector $\tilde{\mu}^{t+1/2}$ can be equivalently written as

$$\tilde{\mu}_{i,u}^{t+1/2} = \frac{\tilde{\mu}_{i,u}^t \cdot \exp(\beta \Delta_{i,u}^{t+1})}{\sum_{i',u'} \tilde{\mu}_{i',u'}^t \cdot \exp(\beta \Delta_{i',u'}^{t+1})}, \quad \forall i \in 1, \ldots, D, u \in 1, \ldots, U.$$

Recall that $\check{v} = \arg\min_{\tilde{v} \in \mathcal{V}} \|\Phi \tilde{v} - v^*\|_\infty$ and $\check{\mu} = \arg\min_{\tilde{\mu} \in \mathcal{U}} \|\Phi \tilde{\mu} \Psi^\top - \mu^*\|_{1,1}$. We obtain that

$$D_{KL}(\check{\mu} \| \tilde{\mu}^{t+1/2}) - D_{KL}(\check{\mu} \| \tilde{\mu}^t) = \sum_{i=1}^D \sum_{u=1}^U \check{\mu}_{i,u} \log \frac{\check{\mu}_{i,u}}{\tilde{\mu}_{i,u}^{t+1/2}} - \sum_{i=1}^D \sum_{u=1}^U \check{\mu}_{i,u} \log \frac{\check{\mu}_{i,u}}{\tilde{\mu}_{i,u}^t}$$

$$= \sum_{i=1}^D \sum_{u=1}^U \check{\mu}_{i,u} \log \frac{\tilde{\mu}_{i,u}^t}{\tilde{\mu}_{i,u}^{t+1/2}}$$

$$= \sum_{i=1}^D \sum_{u=1}^U \check{\mu}_{i,u} \log \frac{Z}{\exp(\beta \Delta_{i,u}^{t+1})}$$

$$= \log Z - \beta \sum_{i=1}^D \sum_{u=1}^U \check{\mu}_{i,u} \Delta_{i,u}^{t+1},$$

where we let $Z = \sum_{i=1}^D \sum_{u=1}^U \tilde{\mu}_{i,u}^t \cdot \exp(\beta \Delta_{i,u}^{t+1})$. According to the definition of $\mathcal{V}$, we have $|\Phi_{s*}\tilde{v}^t| \leq 2t_{mix}$ for all state $s$. Combining with our choice of $M = 4t_{mix} + 1$, we have $\Delta_{i,u}^{t+1} \leq 0$ for all $i = 1, \ldots, D$ and $u = 1, \ldots, U$. Consequently, applying the inequalities $e^x \leq 1 + x + \frac{1}{2}x^2$ for all $x \leq 0$ and $\log(1+x) \leq x$ for all $x > -1$, we have

$$\log Z = \log \sum_{i=1}^D \sum_{u=1}^U \tilde{\mu}_{i,u}^t \cdot \exp(\beta \Delta_{i,u}^{t+1}) \leq \log \sum_{i=1}^D \sum_{u=1}^U \tilde{\mu}_{i,u}^t \left(1 + \beta \Delta_{i,u}^{t+1} + \frac{\beta^2}{2}(\Delta_{i,u}^{t+1})^2\right)$$

$$= \log \left(1 + \beta \sum_{i=1}^D \sum_{u=1}^U \tilde{\mu}_{i,u}^t \Delta_{i,u}^{t+1} + \frac{\beta^2}{2} \sum_{i=1}^D \sum_{u=1}^U \tilde{\mu}_{i,u}^t (\Delta_{i,u}^{t+1})^2\right)$$

$$\leq \beta \sum_{i=1}^D \sum_{u=1}^U \tilde{\mu}_{i,u}^t \Delta_{i,u}^{t+1} + \frac{\beta^2}{2} \sum_{i=1}^D \sum_{u=1}^U \tilde{\mu}_{i,u}^t (\Delta_{i,u}^{t+1})^2$$

Combining the above results, we have

$$D_{KL}(\check{\mu} \| \tilde{\mu}^{t+1/2}) - D_{KL}(\check{\mu} \| \tilde{\mu}^t) \leq \beta \sum_{i=1}^D \sum_{u=1}^U (\tilde{\mu}_{i,u}^t - \check{\mu}_{i,u}) \Delta_{i,u}^{t+1} + \frac{\beta^2}{2} \sum_{i=1}^D \sum_{u=1}^U \tilde{\mu}_{i,u}^t (\Delta_{i,u}^{t+1})^2. \tag{14}$$

In order to prove Lemma 1, we now show that $\mathbf{E}[\Delta_{i,u}^{t+1} \mid \mathcal{F}_t] = \sum_{a \in \mathcal{A}} \Psi_{a,u} \Phi_{*i}^\top ((P_a - I)\Phi \tilde{v}^t + r_a - M \cdot \mathbf{1}_S)$ and that $\sum_{i=1}^D \sum_{u=1}^U \tilde{\mu}_{i,u}^t \mathbf{E}[(\Delta_{i,u}^{t+1})^2 \mid \mathcal{F}_t] \leq 100 DU t_{mix}^2$. We use $\mathbf{1}_S$ to denote the all one column vector with dimension $S$. Recall that $(i_t, u_t)$ is sampled from $\tilde{\mu}^t$, $s_t$ is sampled from $\phi_{i_t}$, $a_t$ is sampled from $\psi_{u_t}$ and $s'_t$ is sampled from $P_{u_t}(s_t, \cdot)$. Hence, for all $(i, u)$, we have

$$\mathbf{E}[\Delta_{i,u}^{t+1} \mid \mathcal{F}_t] = \tilde{\mu}_{i,u}^t \sum_{a \in \mathcal{A}} \sum_{s \in \mathcal{S}} \sum_{s' \in \mathcal{S}} \Psi_{a,u} \cdot \Phi_{s,i} \cdot P_a(s, s') \cdot \frac{\Phi_{s'*} \tilde{v}^t + r_a(s) - \Phi_{s*} \tilde{v}^t - M}{\tilde{\mu}_{i,u}^t}$$

$$= \sum_{a \in \mathcal{A}} \sum_{s \in \mathcal{S}} \Psi_{a,u} \Phi_{s,i} \left( P_a(s, \cdot) \Phi \tilde{v}^t + r_a(s) - \Phi_{s*} \tilde{v}^t - M \right)$$

$$= \sum_{a \in \mathcal{A}} \Psi_{a,u} \Phi_{*i}^\top \left( P_a \Phi \tilde{v}^t + r_a - \Phi \tilde{v}^t - M \cdot \mathbf{1}_S \right).$$

It remains to prove that $\sum_{i=1}^D \sum_{u=1}^U \tilde{\mu}_{i,u}^t \mathbf{E}[(\Delta_{i,u}^{t+1})^2 \mid \mathcal{F}_t] \leq 100 DU t_{mix}^2$. Expanding the expectation, we have

$$\sum_{i=1}^D \sum_{u=1}^U \tilde{\mu}_{i,u}^t \mathbf{E}[(\Delta_{i,u}^{t+1})^2 \mid \mathcal{F}_t]$$

$$= \sum_{i=1}^D \sum_{u=1}^U \tilde{\mu}_{i,u}^t \sum_{a \in \mathcal{A}} \sum_{s \in \mathcal{S}} \sum_{s' \in \mathcal{S}} \Psi_{a,u} \cdot \tilde{\mu}_{i,u}^t \cdot \Phi_{s,i} \cdot P_a(s, s') \left( \frac{\Phi_{s'*} \tilde{v}^t + r_a(s) - \Phi_{s*} \tilde{v}^t - M}{\tilde{\mu}_{i,u}^t} \right)^2$$

$$= \sum_{i=1}^D \sum_{u=1}^U \sum_{a,s,s'} \Psi_{a,u} \cdot \Phi_{s,i} \cdot P_a(s, s') (\Phi_{s'*} \tilde{v}^t + r_a(s) - \Phi_{s*} \tilde{v}^t - M)$$

$$\leq \sum_{i=1}^D \sum_{u=1}^U \sum_{a,s,s'} \Psi_{a,u} \cdot \Phi_{s,i} \cdot P_a(s, s') (8 t_{mix} + 2)^2$$

$$= DU(8 t_{mix} + 2)^2 \leq 100 DU t_{mix}^2,$$

where the first inequality uses the relation that $|\Phi_{s'*} \tilde{v}^t + r_a(s) - \Phi_{s*} \tilde{v}^t - M| \leq 8 t_{mix} + 2$, the third equality is due to that $\sum_{a,s,s'} \Psi_{a,u} \cdot \Phi_{s,i} \cdot P_a(s, s') = 1$ and the last inequality is because $t_{mix} \geq 1$. Substituting the above abounds in equation (14), we obtain that

$$\mathbf{E}[D_{KL}(\check{\mu} \| \tilde{\mu}^{t+1/2}) \mid \mathcal{F}_t] - D_{KL}(\check{\mu} \| \tilde{\mu}^t)$$

$$\leq \beta \sum_{a \in \mathcal{A}} \sum_{i=1}^D \sum_{u=1}^U (\tilde{\mu}_{i,u}^t - \check{\mu}_{i,u}) \Psi_{a,u} \Phi_{*i}^\top ((P_a - I) \Phi \tilde{v}^t + r_a - M \cdot \mathbf{1}_S) + \frac{\beta^2}{2} \cdot 100 DU t_{mix}^2$$

$$\leq \beta \sum_{a \in \mathcal{A}} \Psi_{a*} (\tilde{\mu}^t - \check{\mu})^\top \Phi^\top ((P_a - I) \Phi \tilde{v}^t + r_a) + 50 \beta^2 DU t_{mix}^2,$$

where the last inequality is due to that

$$\sum_{a \in \mathcal{A}} \Psi_{a*} (\tilde{\mu}^t)^\top \Phi^\top \mathbf{1}_S = \sum_{a \in \mathcal{A}} \Psi_{a*} (\check{\mu})^\top \Phi^\top \mathbf{1}_S = 1.$$

Recall that $\tilde{\mu}^{t+1} = \Pi_{\mathcal{U}, KL} (\tilde{\mu}^{t+1/2}) = \operatorname{argmin}_{\mu' \in \mathcal{U}} D_{KL}(\mu' \| \tilde{\mu}^{t+1/2})$ and $\mathcal{U}$ is a convex set. By the property of information projection with regard to KL divergence (see Cover and Thomas (2012) Theorem 11.6.1 on page 367), we have

$$\mathbf{E}[D_{KL}(\check{\mu} \| \tilde{\mu}^{t+1}) \mid \mathcal{F}_t] \leq \mathbf{E}[D_{KL}(\check{\mu} \| \tilde{\mu}^{t+1/2}) \mid \mathcal{F}_t].$$

Combining the above inequalities, we conclude that

$$\mathbf{E}[D_{KL}(\check{\mu} \| \tilde{\mu}^{t+1}) \mid \mathcal{F}_t] - D_{KL}(\check{\mu} \| \tilde{\mu}^t) \leq \mathbf{E}[D_{KL}(\check{\mu} \| \tilde{\mu}^{t+1/2}) \mid \mathcal{F}_t] - D_{KL}(\check{\mu} \| \tilde{\mu}^t)$$

$$\leq \beta \sum_{a \in \mathcal{A}} \Psi_{a*} (\tilde{\mu}^t - \check{\mu})^\top \Phi^\top ((P_a - I) \Phi \tilde{v}^t + r_a) + 50 \beta^2 DU t_{mix}^2,$$

Finally, observe that

$$D_{KL}(\check{\mu}\|\tilde{\mu}^1) = \sum_{i=1}^{D}\sum_{u=1}^{U} \check{\mu}_{i,u} \log \frac{\check{\mu}_{i,u}}{1/(DU)} = \sum_{i=1}^{D}\sum_{u=1}^{U} \check{\mu}_{i,u} \log(DU) + \sum_{i=1}^{D}\sum_{u=1}^{U} \check{\mu}_{i,u} \log(\check{\mu}_{i,u}) \leq \log(DU),$$

where the last inequality is due to that $\check{\mu}_{i,u} \leq 1$ and thus $\log(\check{\mu}_{i,u}) \leq 0$ for all $i, a$. To this point, we complete the proof of Lemma 1. □

## B Proof of Lemma 2

*Proof.* Let $(i_t, u_t, s_t, a_t, s'_t, r_t)$ be the sample at iteration $t$. Throughout the proof, we use the shorthand $\Delta^{t+1} \triangleq \Phi^\top_{s'_t *} - \Phi^\top_{s_t *}$. According to the update of Algorithm 1, we have $\tilde{v}^{t+1} = \Pi_{\mathcal{V}}(\tilde{v}^t - \alpha \Delta^{t+1})$. By using the nonexpansize property of $\Pi_{\mathcal{V}}$, we obtain that

$$\begin{aligned}
\mathbf{E}\left[\|\tilde{v}^{t+1} - \check{v}\|_2^2 \mid \mathcal{F}_t\right] &= \mathbf{E}\left[\|\Pi_{\mathcal{V}}(\tilde{v}^t - \alpha \Delta^{t+1}) - \check{v}\|_2^2 \mid \mathcal{F}_t\right] \leq \mathbf{E}\left[\|\tilde{v}^t - \alpha \Delta^{t+1} - \check{v}\|_2^2 \mid \mathcal{F}_t\right] \\
&= \|\tilde{v}^t - \check{v}\|_2^2 - 2\alpha \mathbf{E}[(\Delta^{t+1})^\top \mid \mathcal{F}_t](\tilde{v}^t - \check{v}) + \alpha^2 \mathbf{E}[\|\Delta^{t+1}\|_2^2 \mid \mathcal{F}_t].
\end{aligned} \quad (15)$$

Recall that $(i_t, u_t)$ is sampled from $\tilde{\mu}^t$, $a_t$ is sampled from $\psi_{u_t}$, $s_t$ is sampled from $\phi_{i_t}$ and $s'_t$ is sampled from $P_{a_t}(s_t, \cdot)$. We can expand the expectation of $\mathbf{E}[(\Delta^{t+1}) \mid \mathcal{F}_t]$ to obtain that

$$\begin{aligned}
\mathbf{E}[(\Delta^{t+1})^\top \mid \mathcal{F}_t] &= \sum_{a \in \mathcal{A}}\sum_{i=1}^{D}\sum_{u=1}^{U}\sum_{s \in \mathcal{S}}\sum_{s' \in \mathcal{S}} \Psi_{a,u} \tilde{\mu}^t_{i,u} \Phi_{s,i} P_a(s, s')(\Phi_{s'*} - \Phi_{s*}) \\
&= \sum_{a \in \mathcal{A}}\sum_{i=1}^{D}\sum_{u=1}^{U}\sum_{s \in \mathcal{S}} \Psi_{a,u} \tilde{\mu}^t_{i,u} \Phi_{s,i}(P_a(s, \cdot)\Phi - \Phi_{s*}) = \sum_{a \in \mathcal{A}}\sum_{i=1}^{D}\sum_{u=1}^{U} \Psi_{a,u} \tilde{\mu}^t_{i,u} \Phi^\top_{*i}(P_a\Phi - \Phi) \\
&= \sum_{a \in \mathcal{A}}\sum_{u=1}^{U} \Psi_{a,u}(\tilde{\mu}^t_{*u})^\top \Phi^\top(P_a\Phi - \Phi) = \sum_{a \in \mathcal{A}} \Psi_{a*}(\tilde{\mu}^t)^\top \Phi^\top(P_a\Phi - \Phi).
\end{aligned}$$

Next we prove that $\mathbf{E}[\|\Delta^{t+1}\|_2^2 \mid \mathcal{F}_t] \leq \|\Phi\|_{2,\infty}^2$. A straightforward calculation yields that

$$\begin{aligned}
\mathbf{E}[\|\Delta^{t+1}\|_2^2 \mid \mathcal{F}_t] &= \sum_{a \in \mathcal{A}}\sum_{i=1}^{D}\sum_{u=1}^{U}\sum_{s \in \mathcal{S}}\sum_{s' \in \mathcal{S}} \Psi_{a,u} \tilde{\mu}^t_{i,u} \Phi_{s,i} P_a(s, s')\|\Phi_{s'*} - \Phi_{s*}\|_2^2 \\
&\leq \sum_{a \in \mathcal{A}}\sum_{i=1}^{D}\sum_{u=1}^{U}\sum_{s \in \mathcal{S}}\sum_{s' \in \mathcal{S}} \Psi_{a,u} \tilde{\mu}^t_{i,u} \Phi_{s,i} P_a(s, s')(2\|\Phi_{s'*}\|_2^2 + 2\|\Phi_{s*}\|_2^2) \\
&\leq \sum_{a \in \mathcal{A}}\sum_{i=1}^{D}\sum_{u=1}^{U}\sum_{s \in \mathcal{S}}\sum_{s' \in \mathcal{S}} \Psi_{a,u} \tilde{\mu}^t_{i,u} \Phi_{s,i} P_a(s, s')(4\|\Phi\|_{2,\infty}^2) = 4\|\Phi\|_{2,\infty}^2,
\end{aligned}$$

where the last equality is due to that $\tilde{\mu}$, $\psi_u$ and $\phi_i$ are distributions and $\sum_{i,u,a,s,s'} \Psi_{a,u} \tilde{\mu}^t_{i,u} \Phi_{s,i} P_a(s, s') = 1$. Substituting the above bounds into equation (15), we get the first part of Lemma 2.

It remains to show that $\|\tilde{v}^1 - \check{v}\|_2^2 = \|\check{v}\|_2^2 \leq \frac{4Dt_{mix}^2 \|\Phi\|_1^2}{\lambda_{\min}^2(\Phi^\top \Phi)}$. We define $v' \triangleq \Phi \check{v}$ to be the projection of $\check{v}$ onto $\mathbb{R}^S$ with regard to $\Phi$. If we multiply $v'$ by $\Phi^\top$, we get $\Phi^\top v' = \Phi^\top \Phi \check{v}$. Hence, by Assumption 1 that $\Phi^\top \Phi$ is invertible, we have

$$\check{v} = (\Phi^\top \Phi)^{-1} \Phi^\top v'$$

By our definition of $\check{v}$ and $\mathcal{V}$, we have $\|v'\|_\infty \leq 2t_{mix}$. Using the relation that $\lambda_{\max}((\Phi^\top \Phi)^{-1}) = \frac{1}{\lambda_{\min}(\Phi^\top \Phi)}$ where $\lambda_{\max}$ and $\lambda_{\min}$ denotes the largest and the smallest eigenvalue, we obtain

$$\begin{aligned}
\|\check{v}\|_2^2 &\leq \|(\Phi^\top \Phi)^{-1}\|_2^2 \|\Phi^\top v'\|_2^2 \leq \frac{1}{\lambda_{\min}^2(\Phi^\top \Phi)} \cdot 4t_{mix}^2 \cdot \|\Phi^\top\|_{1,2}^2 \\
&\leq \frac{4t_{mix}^2}{\lambda_{\min}^2(\Phi^\top \Phi)} \cdot D \cdot \|\Phi^\top\|_{1,\infty}^2 = \frac{4t_{mix}^2 D \|\Phi\|_1^2}{\lambda_{\min}^2(\Phi^\top \Phi)}.
\end{aligned}$$

As a result, we have $\|\check{v}\|_2^2 \le \frac{4t_{mix}^2 D\|\Phi\|_1^2}{\lambda_{\min}^2(\Phi^\top \Phi)}$. Recall that every column of $\Phi$ is a distribution, which gives us $\|\check{v}\|_2^2 \le \frac{4t_{mix}^2 D}{\lambda_{\min}^2(\Phi^\top \Phi)}$. $\square$

## C Proof of Theorem 4

*Proof.* All the norms used in the proof of Theorem 4 are matrix norms. For a matrix $\Phi$ of size $m \times n$, the matrix $p$-norm for $1 \le p \le \infty$ is defined as $\|\Phi\|_p = \max\{\|\Phi v\|_p : v \in \mathbb{R}^n \text{ with } \|v\|_p = 1\}$. Especially, $\|\Phi\|_1$ is the maximum absolute column sum and $\|\Phi\|_\infty$ is the maximum absolute row sum.

We begin by analyzing the behavior of the duality gap in Theorem 2. By some algebra, we can rewrite the LFS of equation (9) as

$$\sum_{a \in \mathcal{A}} r_a^\top \mu_{*a}^* + \frac{1}{T} \sum_{t=1}^T \mathbf{E}\Big[\sum_{a \in \mathcal{A}} ((I - P_a)v^* - r_a)^\top \Phi \tilde{\mu}^t \Psi_{a*}^\top\Big]$$
$$- \underbrace{\frac{1}{T} \sum_{t=1}^T \mathbf{E}\left[\sum_{a \in \mathcal{A}} (\Phi\check{\mu}\Psi_{a*}^\top)^\top (I - P_a)\Phi\tilde{v}^t\right]}_{(i)} + \underbrace{\sum_{a \in \mathcal{A}} (\Phi\check{\mu}\Psi_{a*}^\top - \mu_{*a}^*)^\top r_a}_{(ii)} \qquad (16)$$
$$+ \underbrace{\frac{1}{T} \sum_{t=1}^T \mathbf{E}\Big[\sum_{a \in \mathcal{A}} (\Phi\tilde{\mu}^t \Psi_{a*}^\top)^\top (I - P_a)(\Phi\check{v} - v^*)\Big]}_{(iii)},$$

where $\mu_{*a}^*$ is the $a$-th column of $\mu^*$. Next, we bound (i), (ii), (iii) respectively.

Analysis of (i): Recall that the stationary distribution $\mu^*$ satisfies the condition $\sum_{a \in \mathcal{A}}(\mu_{*a}^*)^\top (I - P_a) = \mathbf{0}_S$. So we can bound (i) by

$$|(\text{i})| \le \left\|\sum_{a \in \mathcal{A}} (\Phi\check{\mu}\Psi_{a*}^\top - \mu_{*a}^*)^\top (I - P_a)\right\|_\infty \left\|\frac{1}{T}\sum_{t=1}^T \mathbf{E}[\Phi\tilde{v}^t]\right\|_\infty$$
$$\le \sum_{a \in \mathcal{A}} \left\|(\Phi\check{\mu}\Psi_{a*}^\top - \mu_{*a}^*)^\top\right\|_\infty (\|I\|_\infty + \|P_a\|_\infty) \cdot 2t_{mix}$$
$$\le 4t_{mix}\|\Phi\check{\mu}\Psi^\top - \mu^*\|_{1,1},$$

where the first inequality is due to that $\|\Phi_1 \Phi_2\|_\infty \le \|\Phi_1\|_\infty \|\Phi_2\|_\infty$ for two matrices $\Phi_1$ and $\Phi_2$, the second inequality is due to that $\|\Phi\tilde{v}^t\|_\infty \le 2t_{mix}$ for all $t$ (see Lemma 1 in Wang (2017)). In the third inequality, we use the fact that the matrix $\infty$-norm of a row vector when viewed as a matrix is the sum of its components. And thus we have $\sum_{a \in \mathcal{A}} \left\|(\Phi\check{\mu}\Psi_{a*}^\top - \mu_{*a}^*)^\top\right\|_\infty = \|\Phi\check{\mu}\Psi^\top - \mu^*\|_{1,1}$.

Analysis of (ii): Using the inequality that $\|\Phi_1 \Phi_2\|_\infty \le \|\Phi_1\|_\infty \|\Phi_2\|_\infty$ for two matrices $\Phi_1, \Phi_2$, we have

$$|(\text{ii})| \le \sum_{a \in \mathcal{A}} \|(\Phi\check{\mu}\Psi_{a*}^\top - \mu_{*a}^*)^\top\|_\infty \|r_a\|_\infty \le \|\Phi\check{\mu}\Psi^\top - \mu^*\|_{1,1},$$

where the last inequality is due to that all the rewards are bounded between 0 and 1.

Analysis of (iii): We note that for any iteration $t$, $\sum_{a \in \mathcal{A}}(\Phi\tilde{\mu}^t \Psi_{a*}^\top)^\top I$ and $\sum_{a \in \mathcal{A}}(\Phi\tilde{\mu}^t \Psi_{a*}^\top)^\top P_a$ are two row vectors that both sum to 1. Recall that the matrix $\infty$-norm of a row vector is the sum of its components. Thus, we have $\|\sum_{a \in \mathcal{A}}(\Phi\tilde{\mu}^t \Psi_{a*}^\top)^\top I - \sum_{a \in \mathcal{A}}(\Phi\tilde{\mu}^t \Psi_{a*}^\top)^\top P_a\|_\infty \le 2$. As a result, we have

$$|(\text{iii})| \le \left\|\frac{1}{T}\sum_{t=1}^T \mathbf{E}\Big[\sum_{a \in \mathcal{A}}(\Phi\tilde{\mu}^t \Psi_{a*}^\top)^\top (I - P_a)\Big]\right\|_\infty \|\Phi\check{v} - v^*\|_\infty$$
$$\le 2\|\Phi\check{v} - v^*\|_\infty,$$

By Theorem 2, we have the relation that $(16) = \tilde{\mathcal{O}}\left(t_{mix}\left(c_\Phi + \sqrt{U\log(DU)}\right)\sqrt{\frac{D}{T}}\right)$. By equation (13), the first two terms of (16) is larger than $\frac{1}{\tau}(\bar{v}^* - \mathbf{E}[\bar{v}^{\hat{\pi}}])$. Combining the above results and the bounds on (i), (ii) and (iii), we draw the conclusion of Theorem 4. □